\documentclass[twoside,11pt]{article}

\usepackage{jmlr2e}

\usepackage{xcolor}
\usepackage[normalem]{ulem}

\ShortHeadings{Bellman: A Toolbox for Model-Based Reinforcement Learning in TensorFlow}{J. McLeod, H. Stoji\'{c}, V. Adam, D. Kim, J. Grau-Moya, P. Vrancx and F. Leibfried}

\firstpageno{1}

\begin{document}

\title{Bellman: A Toolbox for Model-Based \\Reinforcement Learning in TensorFlow}


\author{\name John McLeod \email johnangusmcleod@gmail.com \\
       \addr Secondmind, Cambridge, UK
       \AND
       \name Hrvoje Stoji\'{c} \email hrvoje.stojic@protonmail.com \\
       \addr Secondmind, Cambridge, UK
       \AND
       \name Vincent Adam \email vincent.adam87@gmail.com \\
       \addr Secondmind, Cambridge, UK
       \AND
       \name Dongho Kim \email dongho.kim@gmail.com \\
       \addr Secondmind, Cambridge, UK
       \AND
       \name Jordi Grau-Moya \email jordi.grau.mo@gmail.com \\
       \addr Secondmind, Cambridge, UK
       \AND
       \name Peter Vrancx\thanks{Work done while at Secondmind.} \email peter.vrancx@imec.be \\
       \addr IMEC, Louvain, Belgium 
       \AND
       \name Felix Leibfried \email felix.leibfried@gmail.com \\
       \addr Secondmind, Cambridge, UK}


\maketitle

\begin{abstract}
In the past decade, model-free reinforcement learning (RL) has provided solutions to challenging domains such as robotics. Model-based RL (where agents learn a model of the environment in order to explicitly plan ahead) shows the prospect of being more sample-efficient than model-free methods in terms of agent-environment interactions, because the model enables to extrapolate to unseen situations. In the more recent past, model-based methods have shown superior results compared to model-free methods in some challenging domains with non-linear state transitions. At the same time, it has become apparent that RL is not market-ready yet and that many real-world applications are going to require model-based approaches, because model-free methods are too sample-inefficient and show poor performance in early stages of training. The latter is particularly important in industry, e.g.\ in production systems that directly impact a company's revenue. This demonstrates the necessity for a toolbox to push the boundaries for model-based RL. While there is a plethora of toolboxes for model-free RL, model-based RL has received little attention in terms of toolbox development. Bellman aims to fill this gap and introduces the first thoroughly designed and tested model-based RL toolbox using state-of-the-art software engineering practices. Our modular approach enables to combine a wide range of environment models with generic model-based agent classes that recover state-of-the-art algorithms. We also provide an experiment harness to compare both model-free and model-based agents in a systematic fashion w.r.t.\ user-defined evaluation metrics (e.g.\ cumulative reward). This paves the way for new research directions, e.g.\ investigating uncertainty-aware environment models that are not necessarily neural-network-based, or developing algorithms to solve industrially-motivated benchmarks that share characteristics with real-world problems.
\end{abstract}


\section{Existing Toolboxes for Reinforcement Learning}

In recent years, plenty of toolboxes with state-of-the-art model-free RL algorithms have been developed. We enumerate some of them in the following but do not claim this to be an exhaustive list. Amongst the earliest were ``RL-Glue''~\citep{Tanner2009}, ``rllab''~\citep{Duan2016} and ``keras-rl''~\citep{Plappert2016}, followed by ``Reinforcement Learning Coach''~\citep{Caspi2017}, ``OpenAI Baselines''~\citep{Dhariwal2017} and ``Tensorforce''~\citep{Kuehnle2017}, that were later succeeded by ``Stable Baselines''\citep{Hill2018} and ``RLlib''\citep{Liang2018}. At the time of writing, ``TF-Agents''\citep{Guadarrama2018} in TensorFlow~\citep{Abadi2016} and ``Horizon''~\citep{Gauci2019} in PyTorch~\citep{Paszke2019} appeared to be the most stable and well-maintained toolboxes. 

On the model-based RL side, there are two recent projects that stick out. \cite{Wang2019a} provide a manuscript plus open-sourced code to evaluate various contemporary model-free and model-based RL algorithms on 18 benchmarks, and \cite{Dong2020} provide a software framework called ``Baconian'' (with some algorithms being completed and some still being developed). While the former focuses more on the research aspect and the second more on the software aspect, what both have in common is that algorithms are implemented in a stand-alone fashion in separate classes without modular \& hierarchical object-oriented abstractions and without the flexibility to design new algorithms.

\section{Motivation for a Model-Based Reinforcement Learning Toolbox}

Recently, model-based RL methods have been developed that outperform model-free methods in challenging high-dimensional domains~\citep{Wang2019a}. At the same time, the community became aware of issues that arise when applying RL in practice, and there is an implicit understanding that many real-world problems are going to require model-based approaches~\citep{Dulac2019} for sample-efficiency reasons and because simulators of industrial systems are not available. A second observation is that the literature presents algorithms in a standalone fashion because they are usually developed by different groups, and frameworks are designed in such a way that individual algorithms are implemented in separate classes without inheriting from well-designed object-oriented superclasses that exploit commonalities between algorithms~\citep{Wang2019a,Dong2020}. It is often difficult to tell which precise combination of algorithmic components yields superior results: is it the environment model, the decision-making component, or the choice of optimization objectives? The fact that different papers report results under different environments, reward functions or experiment protocols exacerbates the situation~\citep{Henderson2018}.

The aim of Bellman was to address those problems via identifying commonalities between contemporary model-based RL algorithms and designing a modular software framework where algorithms arise naturally in subclasses and from the composition of well-chosen object-oriented abstractions. This enables the comparison of different algorithms, but importantly, in a controlled setup where the model-based component across different algorithms is the same. Alternatively, different models can be tested simply by changing the environment model object (e.g.\ by replacing a neural net with a Gaussian process). Furthermore, Bellman provides an experiment harness to train and evaluate model-based and -free algorithms systematically and user-conveniently under the same experiment protocol.

\section{Key Features of Bellman and Implementation Details}

\begin{figure}[h!]
\centering
\includegraphics[trim=10 130 0 150,clip,width=\textwidth]{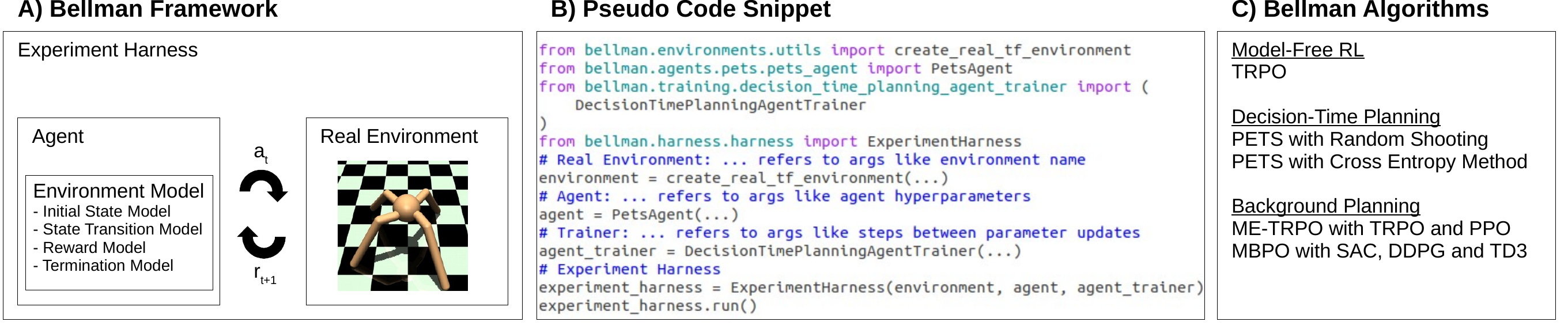}
\caption{Framework overview, pseudo code snippet for PETS, and supported algorithms.}
\label{fig:overview}
\end{figure}

Bellman extends the TensorFlow-based library TF-Agents~\citep{Guadarrama2018} that contains a collection of model-free algorithms and provides three key features on top of it: an environment model API, a model-based RL agent base class that recovers state-of-the-art algorithms via subclasses, and an experiment harness to evaluate both model-based and model-free methods (from TF-Agents) under the same experiment protocol to ensure comparability. See Figure~\ref{fig:overview} for a schematic overview over the framework, a pseudo code snippet for agent assembly and experiment execution, and a list of supported algorithms.

\subsection{Environment Model}

The main idea behind the environment model API is that it mimics a real environment, agents can hence interact with a virtual environment that is learned from data the same way they would interact with the real environment. The environment model comprises individual components for the initial state distribution, state transitions as well as the reward and termination function. At the time of writing, trainable transition models are provided that are neural-network-based and uncertainty-aware, but the software design also enables e.g.\ Gaussian-process-based transitions and to make the other components trainable. 

\subsection{Agent Classes}

The base class for model-based RL contains the environment model components as attributes and a single train method that accepts an extra argument to indicate which component to train. All existing model-based algorithms can be divided into two categories which are covered as subclasses that inherit from our base class design: one for decision-time planning and one for background planning~\citep{Sutton1998}, both of which internally assemble the actual environment model object.

\paragraph{Decision-Time Planners}
 solve a virtual planning problem as determined by the current state of the system and the environment model, the outcome of which is a single action to be executed in the real environment. Bellman provides algorithms that first sample action sequences which are executed in the virtual environment, and then choose the best action sequence proportional to the cumulative trajectory reward. The first action of the best action sequence is then executed. This simple yet versatile type of decision-time planning covers state-of-the-art methods for highly non-linear robotics simulation problems. 
Regarding action-generation, there are two possible scenarios: generating actions either in a state-unconditioned or in a state-conditioned fashion, explained in the following. The former can be achieved e.g.\ with random shooting~\citep{Nagabandi2018} or with the cross entropy method~\citep{Botev2013}, as supported in Bellman through a PETS agent class following \cite{Chua2018}. However, the modular design of Bellman also enables, in principle, state-conditioned action proposals with a virtual policy that maps states to actions. One example for the latter is training a model-free off-policy algorithm on real data but only for virtual execution, as in~\cite{Piche2019}. Another example is to learn a virtual policy through behaviorally cloning the model-based policy, as in~\cite{Wang2019b}. Note that \cite{Piche2019} leverage a specific off-policy algorithm for their purpose, but Bellman is versatile in the sense that it can support any implemented off-policy algorithm.

\paragraph{Background Planners}
cover a large set of algorithms and essentially comprise all model-based RL methods that do planning but are no decision-time planners. In Bellman, however, the term refers to specific algorithms that train model-free algorithms virtually inside of an environment model. Such algorithms are simple yet versatile and recover state-of-the-art approaches for challenging tasks with non-linear transitions. We acknowledge that there are more background planners that are currently not provided in Bellman, e.g.\ model-based value expansion~\citep{Buckman2018} or algorithms focusing on intrinsic motivation~\citep{Leibfried2019b} and robustness~\citep{Grau-Moya2016}.
Bellman provides two subclasses for background planners: one for training on-policy and one for training off-policy model-free algorithms virtually. A specific subclass of the former recovers ME-TRPO~\citep{Kurutach2018} using Bellman's own TRPO~\citep{Schulman2015} implementation. Due to the modular design of Bellman, TRPO can be however conveniently replaced with PPO~\citep{Schulman2017b} from TF-Agents. Another specific model-based off-policy subclass recovers MBPO~\citep{Janner2019} which uses SAC~\citep{Haarnoja2019} internally, but SAC can be replaced by either DDPG~\citep{Lillicrap2016} or TD3~\citep{Fujimoto2018}, as all of those model-free off-policy algorithms are provided by TF-Agents. Note that SLBO~\citep{Liang2018} should also be straightforwardly implementable as it is similar to ME-TRPO and differs only through the model-learning objective and training loop.

\subsection{Experiment Harness}

Bellman provides a single experiment harness class to conduct experiments with both model-free and  model-based agents alike. The experiment harness relies on a flexible configuration for the task, the experiment protocol and the agent. This enables e.g.\ to systematically compare Bellman's PETS, ME-TRPO, MBPO and TRPO implementations against PPO, SAC, DDPG and TD3 from TF-Agents. This systematic comparison is ensured because all experiments can share the same task configuration file (that specifies the real environment) and the same experiment protocol configuration file (that specifies the total number of agent-environment interactions and recording of evaluation metrics), but each experiment can use a different agent configuration file (that specifies agent hyperparameters). Further systematic comparison across model-based algorithms alone can be ensured via a separate environment model configuration file that guarantees that all model-based algorithms share the exact same environment model design and training routine---differences in performance can then be attributed to the decision-making component of each algorithm.

\newpage

\acks{We want to thank Nicolas Durrande, Carl Rasmussen, Juha Sepp\"{a}, Gary Brotman, Peter Bullivant, and everybody else at Secondmind who was involved in the open-sourcing effort.}

\subsection*{The souce code is available at: \url{https://github.com/Bellman-devs/bellman/}}
\bigskip

\appendix
\section{Outlook}

Due to Bellman's modular design, it would be straightforward to combine background planners with extensions of model-free algorithms such as ACKTR~\citep{Wu2017} or MIRACLE~\citep{Leibfried2019} but also with multi-modal policies~\citep{Tang2018,Mazoure2019}. Further work could enable a broader class of model-based algorithms including such that are not planning-based~\citep{Heess2015,Leibfried2018}. On the transition model side, Gaussian-process-based models~\citep{Mattews2017,Leibfried2020,vanderWilk2020,Dutordoir2021} could be readily paired  with our implementations, with the prospect of solving industrially motivated benchmarks~\citep{Hein2017} that would benefit from reliable uncertainty estimates. The latter would effectively extend the seminal work of \cite{Deisenroth2011}, as well as~\cite{Kamthe2018}, to non-linear transition models but, importantly, without the necessity of uni-modal state distribution approximations.

\vskip 0.2in
\bibliography{bellman}

\end{document}